\documentclass[11pt]{article}

\usepackage[final]{acl}

\usepackage{times}
\usepackage{latexsym}

\usepackage[T1]{fontenc}

\usepackage[utf8]{inputenc}

\usepackage{microtype}

\usepackage{inconsolata}

\usepackage{graphicx}
\usepackage{subcaption}  
\usepackage{booktabs}
\usepackage{multirow}
\usepackage{colortbl}
\definecolor{rank1}{HTML}{FFEDED}
\definecolor{rank2}{HTML}{ECF8F1}
\definecolor{rank3}{HTML}{EDEDFF}
\usepackage{amsmath,amssymb}  
\usepackage{dblfloatfix}  
\usepackage{placeins}  
\usepackage[most]{tcolorbox}  
\definecolor{takeawayframe}{HTML}{000099}
\definecolor{takeawayframegreen}{HTML}{009900}
\definecolor{takeawayback}{HTML}{F2F2F2}
\newtcolorbox{takeawaybox}{
  colback=takeawayback,
  colframe=takeawayframe,
  arc=3mm,
  boxrule=0.6pt,
  boxsep=4pt,
  left=10pt,
  right=10pt,
  top=6pt,
  bottom=6pt,
}
\newtcolorbox{takeawayboxgreen}{
  colback=takeawayback,
  colframe=takeawayframegreen,
  arc=3mm,
  boxrule=0.6pt,
  boxsep=4pt,
  left=10pt,
  right=10pt,
  top=6pt,
  bottom=6pt,
}
\newtcolorbox{promptbox}[1][Prompt]{
  enhanced,
  colback=gray!5,
  colframe=black,
  colbacktitle=black,
  coltitle=white,
  fonttitle=\bfseries,
  title={#1},
  arc=1mm,
  boxrule=0.6pt,
  boxsep=4pt,
  left=8pt, right=8pt, top=4pt, bottom=4pt,
}

\title{Where to Look: Can Foundation Models Reach a Target Viewpoint Through Active Exploration?}

\author{
  Liyang Li\textsuperscript{*},
  Muzhi Zhu\textsuperscript{*},
  Zhiyue Zhao,
  Hengyu Zhao,
  Ke Liu \\
  \bfseries Linhao Zhong,
  Hao Chen,
  Chunhua Shen\textsuperscript{\dag} \\
  \mdseries Zhejiang University \\
  \mdseries {\small \textsuperscript{*}Equal contribution \quad \textsuperscript{\dag}Corresponding author}
}

\begin{document}

\makeatletter
\let\ASorigmaketitle\@maketitle
\renewcommand{\@maketitle}{%
  \ASorigmaketitle
  \begin{center}
    \includegraphics[width=\textwidth]{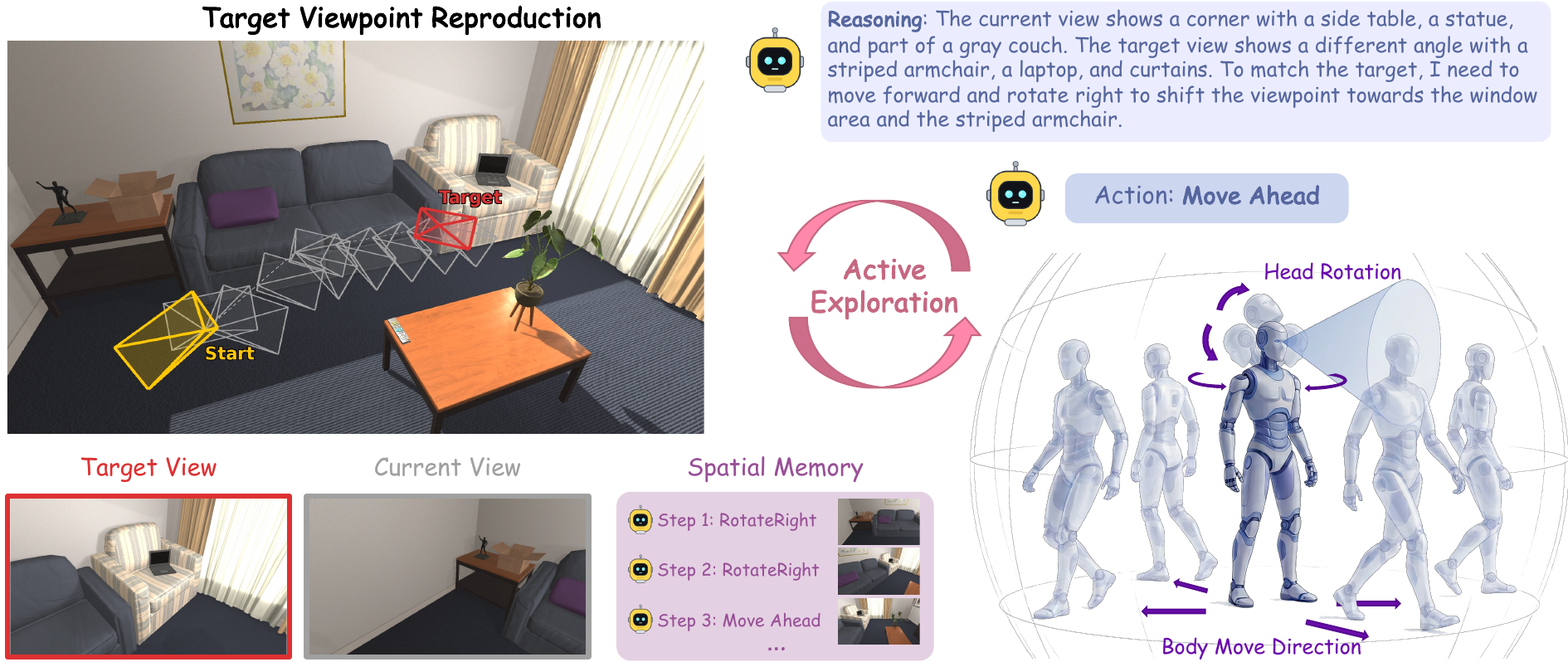}
    \captionof{figure}{\textbf{Target Viewpoint Reproduction (TVR).} Can a foundation model actively reproduce a target viewpoint in 3D, closing the perception--reasoning--action loop through body translation and head rotation?}
    \label{fig:teaser}
  \end{center}
  \vskip 1em
}
\makeatother

\maketitle

\begin{abstract}

Humans can reproduce the viewpoint specified by a target image through active head and body motion, yet spatial intelligence in foundation models has largely been studied as passive understanding of pre-collected observations.
We introduce \textbf{Target Viewpoint Reproduction (TVR)}---an active task where an agent adjusts its viewpoint in a 3D environment until its observation matches a given target image---and \textbf{TVRBench}, an indoor-simulation benchmark spanning scene scale and target-view visual richness.
TVR is far from solved: on the evaluation split, the strongest open-source and closed-source models reach only $7.8\%$ and $12.0\%$ success.
Fine-grained analysis identifies two consistent bottlenecks: off-the-shelf models struggle with multi-turn visual history, and performance drops sharply when viewpoint reproduction requires body translation rather than in-place rotation, exposing a gap in mapping spatial discrepancies to embodied movement.
To study reducing this gap, we build a unified TVR post-training framework covering expert-trajectory SFT, rationale-supervised CoT-SFT, offline Single-turn GRPO, and on-policy Multi-turn GRPO from live simulator rollouts.
Visual-action SFT supplies the main gain, raising a 9B open-source model to $50.8\%$ success; Multi-turn GRPO provides targeted multi-room refinement and reaches $51.4\%$ overall, while CoT supervision and Single-turn GRPO degrade closed-loop performance.
These results establish TVRBench as a testbed for measuring and training foundation models that actively perceive and act in 3D environments. Our code, data, and models are available at \url{https://github.com/aim-uofa/TVRBench}.

\end{abstract}

\section{Introduction}

Reproducing the viewpoint from a single target image is a basic form of active spatial intelligence.
The agent must compare the target with its egocentric view, infer the viewpoint gap, map it to body translation, rotation, and head motion, update spatial belief from new observations, and decide when the match is accurate enough to stop.
Humans do this naturally: instead of passively matching static content, we move in 3D, gather visual evidence, and refine actions through a closed perception--action loop.

Recent spatial-intelligence research on foundation models, especially MLLMs, has introduced diverse tasks and benchmarks for relative position, directional relations, 3D layouts, and cross-view reasoning~\cite{chen2024spatialvlm, cheng2024spatialrgpt, hong20233d, yang2025mmsi}.
Yet most assume visual observations are given in advance, as a static image, multi-view inputs, or a prerecorded video, and thus ask only what is where, not where should I move and look next.
Active-exploration tasks such as ImageNav~\cite{zhu2017target, krantz2022instance} move closer to embodied spatial intelligence, but typically evaluate whether agents reach a target region rather than whether their final egocentric observation reproduces the goal image.

This raises a central question:
can foundation models infer current-to-target spatial relations,
map them to embodied actions,
and reproduce target views through active exploration?
We introduce \textbf{Target Viewpoint Reproduction (TVR)},
where an agent receives a target image and initial observation
in a 3D environment,
then acts until its observation matches the target.
TVR is \emph{active},
gathering new observations in a closed perception--action loop,
and evaluates \emph{explicit viewpoint control}:
the agent must reproduce the target viewpoint
rather than merely reach a region.
We instantiate TVR in indoor simulation as \textbf{TVRBench},
covering single-room and multi-room scenes
with diagnostics for exploration efficiency,
spatial memory, and perception-to-action mapping.

Across the open- and closed-source MLLMs we evaluate,
TVRBench shows TVR remains far from solved:
the strongest open-source model reaches $7.8\%$ success
and the strongest closed-source model $12.0\%$,
versus $93\%$ human performance
on a 100-task subset.
Fine-grained analysis finds two bottlenecks.
First, off-the-shelf models struggle with multi-turn visual history:
every open-source model performs better with an action-only recap
than full visual-action memory
(mean gap $+3.8$\,pp).
Second, performance drops when viewpoint reproduction requires body translation
rather than in-place rotation,
suggesting the main difficulty is mapping spatial discrepancies
to embodied movement,
beyond static visual recognition.

We build a unified TVR post-training framework
to target these bottlenecks,
covering expert-trajectory SFT~\cite{kim2024openvla}, CoT-SFT,
offline Single-turn GRPO~\cite{liao2025improved},
and on-policy Multi-turn GRPO from live rollouts~\cite{zeng2024poliformer}.
It compares models/training paradigms
for closed-loop active perception.
Visual-action SFT gives the main gain,
raising Qwen3.5-9B to $50.8\%$ without CoT.
Multi-turn GRPO refines VA-SFT to $51.4\%$,
mainly on multi-room tasks where SFT is weakest.
In contrast, CoT supervision and Single-turn GRPO reduce success,
suggesting per-step rationales or action matching
may not transfer to embodied multi-step control.

Our main contributions are as follows:
\begin{itemize}
    \item We introduce \textbf{Target Viewpoint Reproduction (TVR)}, a closed-loop target-viewpoint reproduction task, and \textbf{TVRBench}, an indoor-simulation benchmark with protocols diagnosing exploration efficiency, spatial memory, and perception-to-action mapping.

    \item We benchmark open- and closed-source foundation models on TVRBench 
    and identify two consistent bottlenecks: exploiting multi-turn visual history and mapping spatial discrepancies to body translation.

    \item We develop a unified TVR post-training framework for comparing expert-trajectory SFT, CoT-SFT, and single-/multi-turn GRPO in closed-loop environments.

    \item Using this framework, we show that visual-action SFT supplies the main improvement ($50.8\%$) and Multi-turn GRPO provides targeted multi-room refinement ($51.4\%$ overall), while CoT supervision and Single-turn GRPO degrade closed-loop performance.
\end{itemize}

\section{Related Work}

\subsection{Spatial Intelligence}

Early foundation-model work on spatial intelligence addressed static inputs: from text-image pairs or single visual observations, models answer questions about relative position, orientation, directional relations, topology, or 3D layout~\cite{Johnson_2017_CVPR, liu2023visual, wang2024picture, chen2024spatialvlm, cheng2024spatialrgpt, li2025seeground}.
Later work extended this to multi-view settings for cross-view matching, spatial-relation inference, and local-to-global scene-structure understanding~\cite{hong20233d, yeh2026seeing, yin2025spatial, xu2025multi, yang2025mmsi}, and videos, where continuous observations enable spatial updating and temporal reasoning~\cite{yang2025thinking, zhou2025vlm4d}.
Another line grounds spatial reasoning in embodied agents via embodied question answering and affordance prediction~\cite{ma2022sqa3d, majumdar2024openeqa, zhou2024navgpt, yuan2024robopoint}.

Across settings, however, visual observations are typically pre-collected, not acquired through exploration: the model is asked only ``what is where,'' not ``where should I look next.''
\subsection{Active Embodied Reasoning}

Visual navigation dominates active embodied tasks.
Goals are specified by an object class (ObjectNav~\cite{batra2020objectnav}),
a goal image (ImageNav~\cite{zhu2017target, krantz2022instance}),
or a natural-language instruction (VLN~\cite{anderson2018vision, zhou2024navgpt}).
Across settings, success measures whether the agent's \emph{position}
reaches the target region or fulfills the instruction,
rather than whether its final observation reproduces a target viewpoint.
Even ImageNav uses a goal image but scores proximity,
not exact visual match.

Recent work investigates active spatial reasoning with foundation models~\cite{yang2025thinking, yin2025spatial, zhu2025move, zhang2026theory, zhu2025active}, often using simplified action spaces such as teleportation or restricted agent positions.
Concurrent humanoid visual search work~\cite{yu2025thinking} studies head-rotation-only object and path search over $360^{\circ}$ panoramas, while visually grounded active view selection~\cite{koo2025toward} selects informative next views without reproducing a specific target.
\citet{hong2026esi} introduce ESI-Bench, a broad embodied-spatial-intelligence benchmark with ten OmniGibson task categories, and \citet{sakamoto2026e3vs} propose E3VS-Bench for active VQA in 3DGS scenes.

TVR differs from these settings along two axes: (i) success is defined by an explicit viewpoint match---the agent's observation must reproduce the viewpoint of a given target image---rather than reaching a position, identifying an object, or completing an instruction; and (ii) the action space spans both body movement and head rotation, without teleportation, fixed positions, or restriction to a single action modality, demanding coordination.

\subsection{Post-Training for Vision-Language and Embodied Tasks}

Recent work applies post-training to spatial reasoning vision-language models, achieving substantial gains on static spatial benchmarks.
Large-scale SFT on simulator-generated spatial QA establishes the supervised paradigm~\cite{ray2024sat}, while pure GRPO lifts a small VLM past proprietary baselines on video spatial QA~\cite{liao2025improved}, and a two-stage SFT-then-GRPO has emerged as a dominant recipe~\cite{wu2026spatial}.
These methods, however, target static spatial QA from pre-collected inputs rather than closed-loop active perception.

Vision-Language-Action (VLA) models extend pretrained VLMs to robotic control through supervised demonstrations on robot data~\cite{kim2024openvla, black2024pi_0}, framing embodied control as action-token sequence prediction.
Concurrent work on transformer-based on-policy reinforcement learning shows scaling RL produces strong embodied navigators~\cite{zeng2024poliformer}.

Our experiments suggest a mismatch between this per-step paradigm and TVR's active multi-step structure: per-step Single-turn GRPO regresses below its SFT initialisation, while trajectory-level Multi-turn GRPO over live rollouts is required to refine rather than overwrite the supervised priors.

\section{Target Viewpoint Reproduction and TVRBench}

\begin{figure*}[t]
\centering
\includegraphics[width=0.97\textwidth]{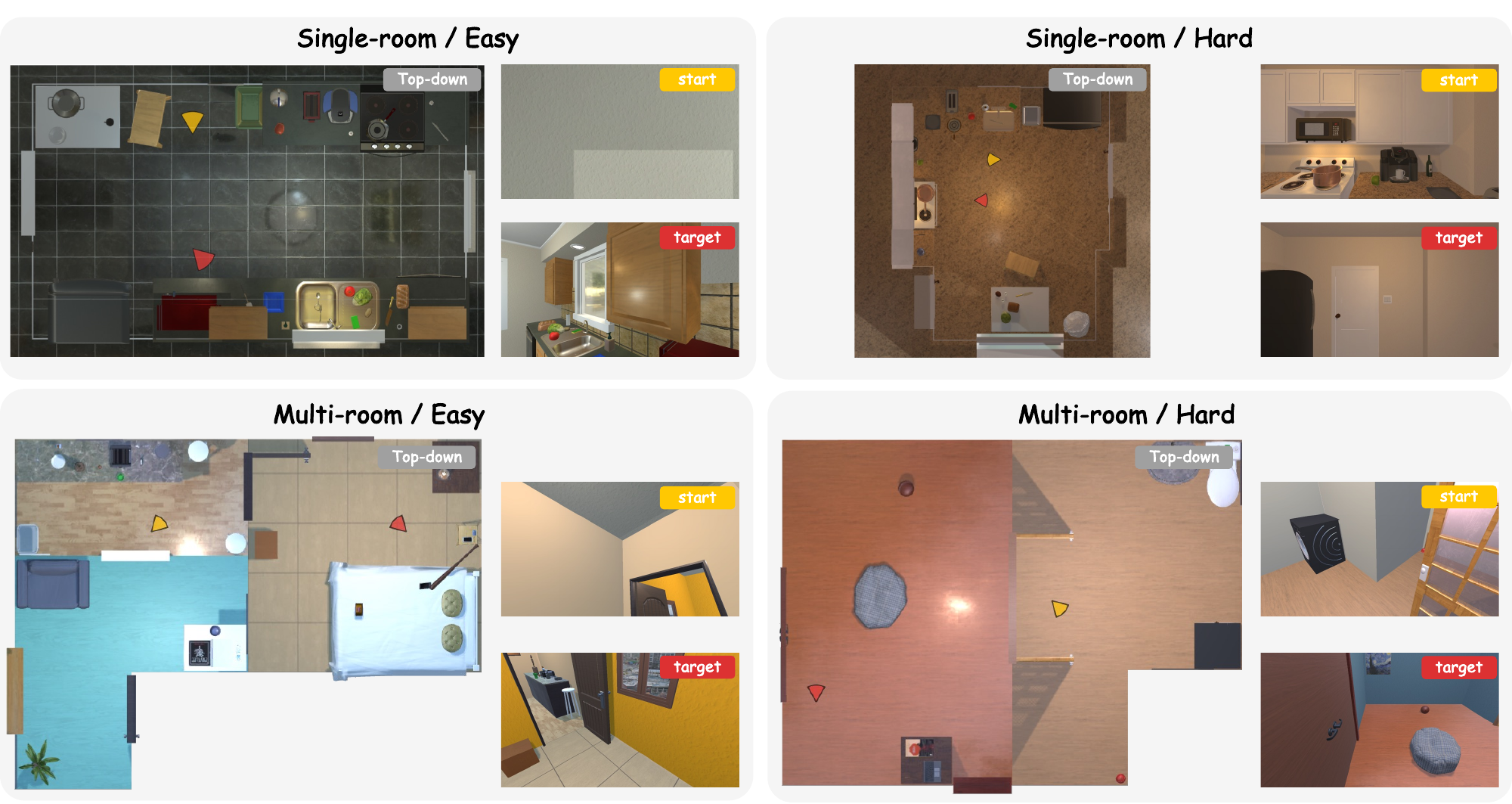}
\caption{\textbf{TVRBench Task Structure} A $2{\times}2$ task design crossing scene scale and target-view visual richness. Each category shows one representative task: an orthographic top-down with start (yellow) and target (red) poses, and first-person views at both poses. We label the four categories Single-easy, Single-hard, Multi-easy, Multi-hard.}
\label{fig:tvrbench}
\end{figure*}

Active spatial intelligence involves more than recognizing what is visible; it also requires choosing where to look next and moving to obtain that view. We study this ability through a task in which success depends on viewpoint recovery, without language grounding as a confounding factor.

\subsection{The TVR Task}

In \textbf{Target Viewpoint Reproduction (TVR)}, an agent operates in a 3D indoor environment and is given a single target image $I^{\star}$ rendered from a viewpoint in the same scene.
At each timestep, the agent observes the current first-person image $I_t$ and selects one action.
The episode ends when the agent selects the Stop action or reaches the step limit for the task.
The agent succeeds only if its final pose exactly matches the viewpoint of $I^{\star}$.

\paragraph{State and action space.}
The agent state $s_t = (x_t, z_t, \theta_t, \phi_t)$ comprises the ground-plane position $(x_t, z_t)$, body yaw $\theta_t$, and camera horizon (head pitch) $\phi_t$.
At each step, the agent selects one of nine discrete agent-centric actions.
MoveAhead, MoveBack, MoveLeft, and MoveRight translate the body by $0.25$\,m.
RotateLeft and RotateRight rotate the body by $45^{\circ}$; LookUp and LookDown shift the camera horizon by $30^{\circ}$.
Stop signals task completion and ends the episode.

\paragraph{Observation and termination.}
At each step, the agent observes only the first-person RGB image $I_t$ rendered from $s_t$, with no privileged access to its pose, the target pose, or a scene map.
An episode ends when the agent issues Stop or reaches the task step limit. The limit is $30$ steps for single-room and $40$ steps for multi-room tasks, as multi-room tasks typically require longer routes.
\paragraph{Success criterion.}
Because action steps and target poses share the same discrete pose grid, the agent can reach the target pose exactly.
An episode succeeds if and only if the agent issues Stop and its final pose $s_T$ is identical to the target pose $s^{\star}$:
\[
s_T = s^{\star}.
\]
Thus, the final observation must exactly match the viewpoint of $I^{\star}$, not merely approximately.
Success is evaluated on the same $0.25\,\mathrm{m}$ pose grid used by the action space. At this resolution, adjacent poses produce distinguishable observations, so exact matching appropriately tests viewpoint identity.
\begin{table*}[!t]
\centering
\small
\setlength{\tabcolsep}{5pt}
\setlength{\fboxsep}{2pt}
\caption{\textbf{Foundation model evaluation on TVRBench.} Success rate (\%) and diagnostics on the test split (S-e/S-h: single-room easy/hard; M-e/M-h: multi-room easy/hard); top-3 per column: \colorbox{rank1}{red}, \colorbox{rank2}{green}, \colorbox{rank3}{blue}.}
\label{tab:fm_eval}
\begin{tabular}{lcccccccccc}
\toprule
\textbf{Model} & \textbf{Overall}$\uparrow$ & \textbf{S-e}$\uparrow$ & \textbf{S-h}$\uparrow$ & \textbf{M-e}$\uparrow$ & \textbf{M-h}$\uparrow$ & \textbf{Steps} & \textbf{F-stop}$\downarrow$ & $\boldsymbol{|\Delta p|}\downarrow$ & $\boldsymbol{|\Delta\theta|}\downarrow$ & $\boldsymbol{|\Delta\phi|}\downarrow$ \\
\midrule
Qwen3.5-9B (VA)    & $0.0$    & $0.0$  & $0.0$  & $0.0$  & $0.0$  & $34.3$ & $100.0$ & $2.05$       & $86.8$            & $15.8$          \\
Qwen3.5-9B (AO)    & $2.8$    & $6.4$  & $4.8$  & $0.0$  & $0.0$  & $29.9$ & $89.6$  & $1.94$       & $67.8$            & $12.8$          \\
Qwen3.5-27B (VA)   & $3.2$    & $5.6$  & $4.0$  & $0.8$  & $2.4$  & $31.2$ & $82.0$  & $1.71$       & $52.6$            & $9.8$           \\
Qwen3.5-27B (AO)   & \cellcolor{rank3}$7.8$    & $14.4$ & \cellcolor{rank2}$13.6$ & $0.0$  & \cellcolor{rank3}$3.2$  & $28.3$ & $75.9$  & $1.57$       & \cellcolor{rank3}$49.2$            & \cellcolor{rank2}$6.6$           \\
Qwen3.5-35B-A3B (VA) & $0.8$  & $2.4$  & $0.8$  & $0.0$  & $0.0$  & $33.8$ & $84.0$  & $2.03$       & $80.4$            & $10.6$          \\
Qwen3.5-35B-A3B (AO) & $3.8$  & $8.0$  & $5.6$  & $0.0$  & $1.6$  & $33.1$ & $67.2$  & $2.00$       & $76.0$            & $9.8$           \\
Qwen3.6-27B (VA)   & $3.2$    & $8.8$  & $3.2$  & $0.8$  & $0.0$  & $30.9$ & $82.6$  & $1.94$       & $57.0$            & $12.9$          \\
Qwen3.6-27B (AO)   & $7.0$    & $12.8$ & \cellcolor{rank3}$9.6$  & \cellcolor{rank3}$1.6$  & \cellcolor{rank2}$4.0$  & $27.2$ & $80.8$  & $1.66$       & $53.7$            & $7.6$           \\
Qwen3.6-35B-A3B (VA) & $0.2$  & $0.0$  & $0.8$  & $0.0$  & $0.0$  & $34.1$ & $94.7$  & $2.10$       & $74.6$            & $14.7$          \\
Qwen3.6-35B-A3B (AO) & $4.8$  & $9.6$  & $8.0$  & $0.8$  & $0.8$  & $28.8$ & $84.5$  & $1.80$       & $70.6$            & $9.3$           \\
\midrule
GPT-4o (VA)        & $2.8$    & $5.6$  & $4.0$  & \cellcolor{rank3}$1.6$  & $0.0$  & $31.3$ & $87.3$  & $1.68$       & $61.1$            & $11.9$          \\
GPT-4o (AO)        & $5.2$    & $13.6$ & $5.6$  & $0.8$  & $0.8$  & $33.6$ & \cellcolor{rank3}$49.0$  & $1.60$       & $52.3$            & $9.2$           \\
GPT-5 (VA)         & \cellcolor{rank2}$8.0$    & \cellcolor{rank3}$15.2$ & $8.0$  & \cellcolor{rank1}$\mathbf{3.2}$  & \cellcolor{rank1}$\mathbf{5.6}$  & $33.7$ & \cellcolor{rank2}$27.3$  & \cellcolor{rank1}$\mathbf{1.30}$       & \cellcolor{rank2}$43.2$            & \cellcolor{rank1}$\mathbf{4.2}$           \\
GPT-5 (AO)         & \cellcolor{rank2}$8.0$    & \cellcolor{rank1}$\mathbf{22.4}$ & $6.4$  & \cellcolor{rank2}$2.4$  & $0.8$  & $34.0$ & \cellcolor{rank1}$\mathbf{0.0}$   & \cellcolor{rank3}$1.54$       & $56.6$            & \cellcolor{rank3}$7.0$           \\
Gemini-3.1-Pro (AO) & \cellcolor{rank1}$\mathbf{12.0}$   & \cellcolor{rank2}$21.6$ & \cellcolor{rank1}$\mathbf{21.6}$ & $0.8$  & \cellcolor{rank2}$4.0$  & $10.3$ & $87.1$  & \cellcolor{rank2}$1.47$       & \cellcolor{rank1}$\mathbf{41.9}$            & $8.2$           \\
\midrule
\emph{Human}       & \emph{$93.0$}   & \emph{$100.0$}& \emph{$88.0$} & \emph{$88.0$} & \emph{$96.0$} & \emph{$22.8$} & \emph{$2.1$}   & \emph{$0.05$}       & \emph{$0.9$}             & \emph{$0.0$}           \\
\bottomrule
\end{tabular}
\end{table*}

\subsection{The TVRBench Benchmark}
\label{sec:tvrbench}

\paragraph{Design rationale.}
TVRBench separates two difficulty sources in viewpoint reproduction:
scene scale and target-view visual evidence.
Scene scale tests whether agents move beyond local adjustment,
as multi-room cases require traversing rooms
to reach target area.
Target-view evidence determines how images disambiguate the viewpoint:
object-rich views provide landmarks and geometric cues,
whereas sparse views offer fewer anchors.
We stratify by scene scale and target-view visual richness,
with easy/hard tiers for each.
The four equal-sized categories,
Single-easy, Single-hard, Multi-easy, and Multi-hard,
support analysis of movement difficulty and target-view evidence.

\paragraph{Scene sources and sampling.}
TVRBench uses two scene sources:
single-room tasks use iTHOR~\cite{ai2thor},
with $120$ kitchens, living rooms, bedrooms, and bathrooms,
while multi-room tasks use ProcTHOR-10k~\cite{procthor},
with two- or three-room homes separated by physical walls.
We split the $240$ scenes, $120$ per source,
into disjoint SFT, evaluation, and RL-training sets
at a $1{:}2{:}3$ ratio,
excluding evaluation scenes from training.
Per scene, we uniformly sample (start, target) pose pairs
from the reachable grid and filter by visible-object count,
the number of non-structural objects\footnote{Walls, floor, ceiling, and the agent itself are excluded from the count.}
visible from the target view,
and shortest start-to-target action-path length.
Easy tasks require at least $9$ target-visible objects,
while hard tasks allow only $3$--$6$.
Shortest paths span $2$--$8$ action steps in single-room scenes
and $10$--$20$ in multi-room scenes.
The benchmark contains $125$ tasks per category
and $500$ evaluation tasks total.
Representative examples appear in Figure~\ref{fig:tvrbench}.

\paragraph{Memory representations.}
The agent needs a trajectory record to avoid revisits
and judge progress toward $I^{\star}$,
making past-step representation an important design choice.
We use two memory representations throughout experiments.
In action-only memory (AO),
the model receives the current observation $I_t$,
target $I^{\star}$,
and a brief summary of previous actions.
In visual-action memory (VA),
the full past observation-action sequence remains available
in a multi-turn multimodal context.
These representations emphasize trade-offs.
VA tests whether a model can effectively use trajectory visual history,
whereas AO reduces the number of images sent per call.
AO makes rate/context-limited closed-source evaluation cheaper/faster.

\paragraph{Evaluation metrics.}
Beyond the binary success criterion, we report three diagnostic metrics describing how an agent fails. The final pose errors $|\Delta p|, |\Delta \theta|, |\Delta \phi|$ between $s_T$ and $s^{\star}$ quantify remaining distance in failed episodes. The stop rate, the fraction of episodes terminating with Stop, and the false-stop rate, the fraction of Stop actions taken at non-target poses, separate cases where the agent never stops from cases where it stops at an incorrect pose. We report the mean number of steps to termination, which measures exploration efficiency.
\section{Can Foundation Models Reproduce Target Viewpoints?}
\label{sec:fm_eval}
%
\begin{figure*}[t]
\centering
\includegraphics[width=0.97\textwidth]{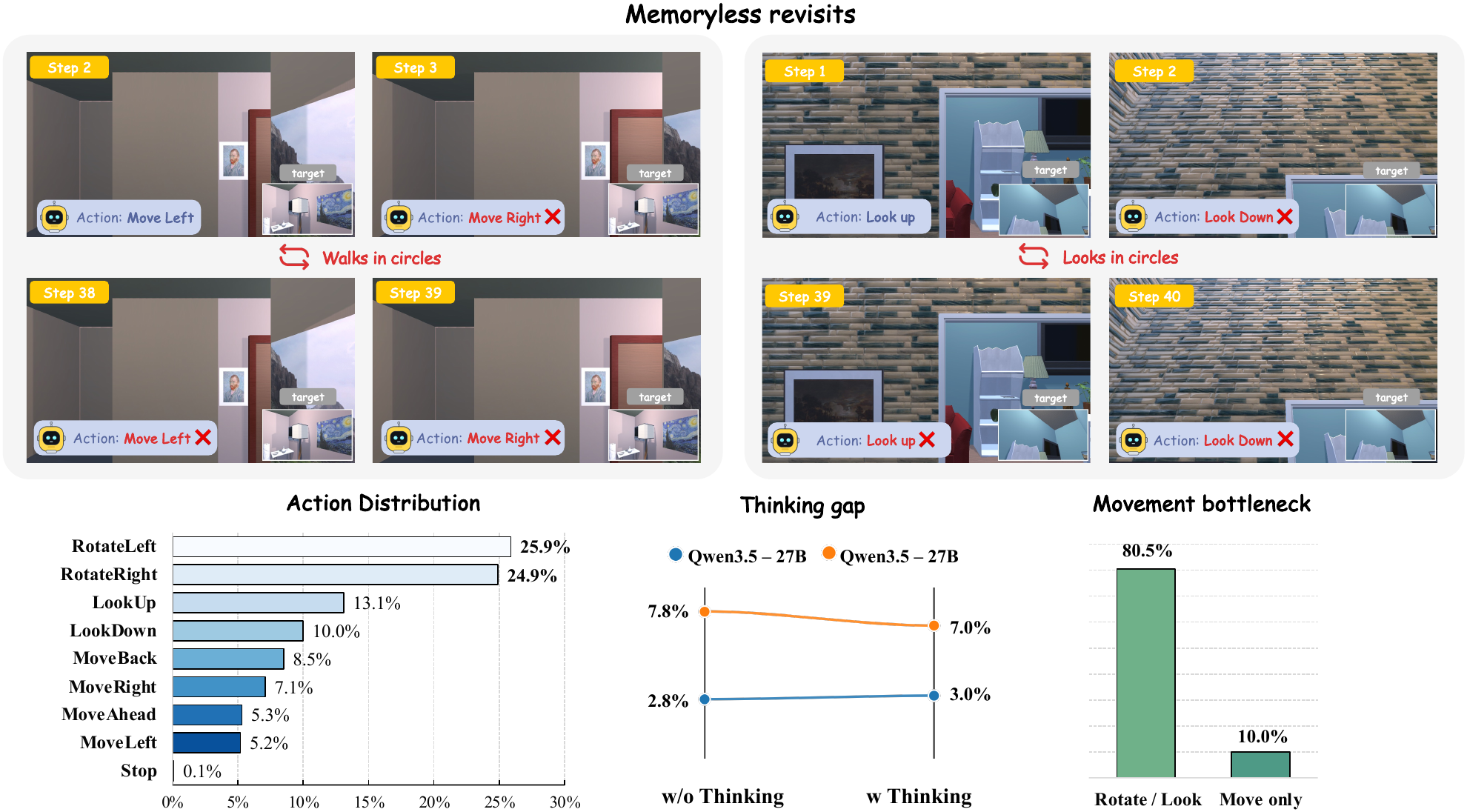}
\caption{\textbf{Why an untrained 9B fails at TVR.} \textbf{Top}: Qwen3.5-9B visits only $3.5$ distinct grid positions per episode and revisits $83\%$ of poses, producing two stable failure modes---\emph{walks in circles} (left) and \emph{looks in loops} (right). \textbf{Bottom-left}: action selection distribution (rotation $50.8\%$, body translation $26.1\%$, Stop $0.1\%$). \textbf{Bottom-middle}: enabling chain-of-thought multiplies tokens per response by $\sim$10$\times$ without changing success rate. \textbf{Bottom-right}: removing body translation lifts the model to $80.5\%$; restricting to it keeps the model at $10.0\%$.}
\label{fig:fm_failures}
\end{figure*}

\paragraph{Models and protocol.}
We benchmark five open-source baselines:
dense Qwen3.5-9B~\cite{qwen3.5},
Qwen3.5-27B, Qwen3.6-27B~\cite{qwen3.6-27b},
and MoE Qwen3.5-35B-A3B
and Qwen3.6-35B-A3B~\cite{qwen36_35b_a3b}.
We also evaluate three closed-source models:
GPT-4o~\cite{hurst2024gpt},
GPT-5~\cite{singh2025openai},
and Gemini-3.1-Pro~\cite{gemini3.1pro}.
All are evaluated on the held-out 500-task split
with step budgets and VA/AO memory settings
defined in Section~\ref{sec:tvrbench}.
Gemini-3.1-Pro is evaluated under AO only,
because VA multi-image inference over the full split
is prohibitively slow.
Open-source models use greedy decoding;
closed-source models use the lowest API-supported temperature.
For reference, we report human performance
from five participants on a balanced 100-task subset,
using the same resolution, action space,
step budget, and success criterion.

\paragraph{Main results.}
Table~\ref{tab:fm_eval} reports success rates by task category
across 15 model-memory configurations.
The best configuration reaches only $12.0\%$ overall success
(Gemini-3.1-Pro, AO),
and no model exceeds $13\%$.
By contrast, humans achieve $93.0\%$
on a balanced 100-task subset.

Scaling brings small gains.
Dense Qwen3.5 improves from $2.8\%$ at 9B to $7.8\%$ at 27B,
while the best closed-source models remain at the $12\%$ ceiling.
Results show a consistent pattern.
Every open-source model performs better under AO than VA,
with a mean gap of $+3.8$\,pp,
suggesting past observations in context can hurt foundation models
not trained for this setting.
When a model invokes Stop,
it usually does so at the wrong pose:
F-stop exceeds $75\%$ for $11$ of $15$ configurations
and reaches $100\%$ for Qwen3.5-9B (VA).
GPT-5 is the exception ($0\%$ AO, $27.3\%$ VA):
when it commits to Stop,
it is usually already at the target pose.
Models rarely terminate on their own:
for $14$ of $15$ rows,
mean episode length is close to the per-task step budget,
so most episodes hit the step limit rather than end with Stop.

\paragraph{Controlled ablation: body translation is a dominant bottleneck.}
To locate failures,
we run two single-room ablations with $200$ tasks each
under restricted action spaces.
In rotate/look,
start/target states share position
and differ only in yaw and head pitch.
In move-only,
they share yaw and pitch
and differ only in position.
Removing body-translation actions raises Qwen3.5-9B
from $2.8\%$ baseline to $80.5\%$,
whereas allowing only body translation keeps it at $10.0\%$
(Figure~\ref{fig:fm_failures}, bottom-right).
Results suggest body-translation control
is a dominant failure mode in TVR,
rather than viewpoint appearance matching alone.

\paragraph{Failure behavior patterns.}
The full benchmark shows three recurring behavioral patterns
consistent with the controlled-ablation result.
Per episode, Qwen3.5-9B chooses $34.3$ actions on average,
yet visits only $3.5$ distinct grid positions
and returns to $83\%$ of its own poses.
Failed trajectories mainly follow two stable patterns:
the agent walks in circles,
moving back and forth between adjacent cells,
or looks in loops,
alternating head pitch while staying put.
Examples appear at the top of Figure~\ref{fig:fm_failures}.

The action distribution points to the same issue.
Among $17{,}159$ actions across the benchmark,
rotations account for $50.8\%$,
body translations only $26.1\%$,
and Stop just $0.1\%$
(Figure~\ref{fig:fm_failures}, bottom-left).
In practice, the model rotates too often
and rarely moves forward or ends the episode.

Enabling Qwen3.5's native thinking mode
does not resolve this behavior.
It increases response tokens by roughly an order of magnitude,
but success remains unchanged
(Figure~\ref{fig:fm_failures}, bottom-middle).

\begin{figure*}[!t]
\centering
\includegraphics[width=0.97\textwidth]{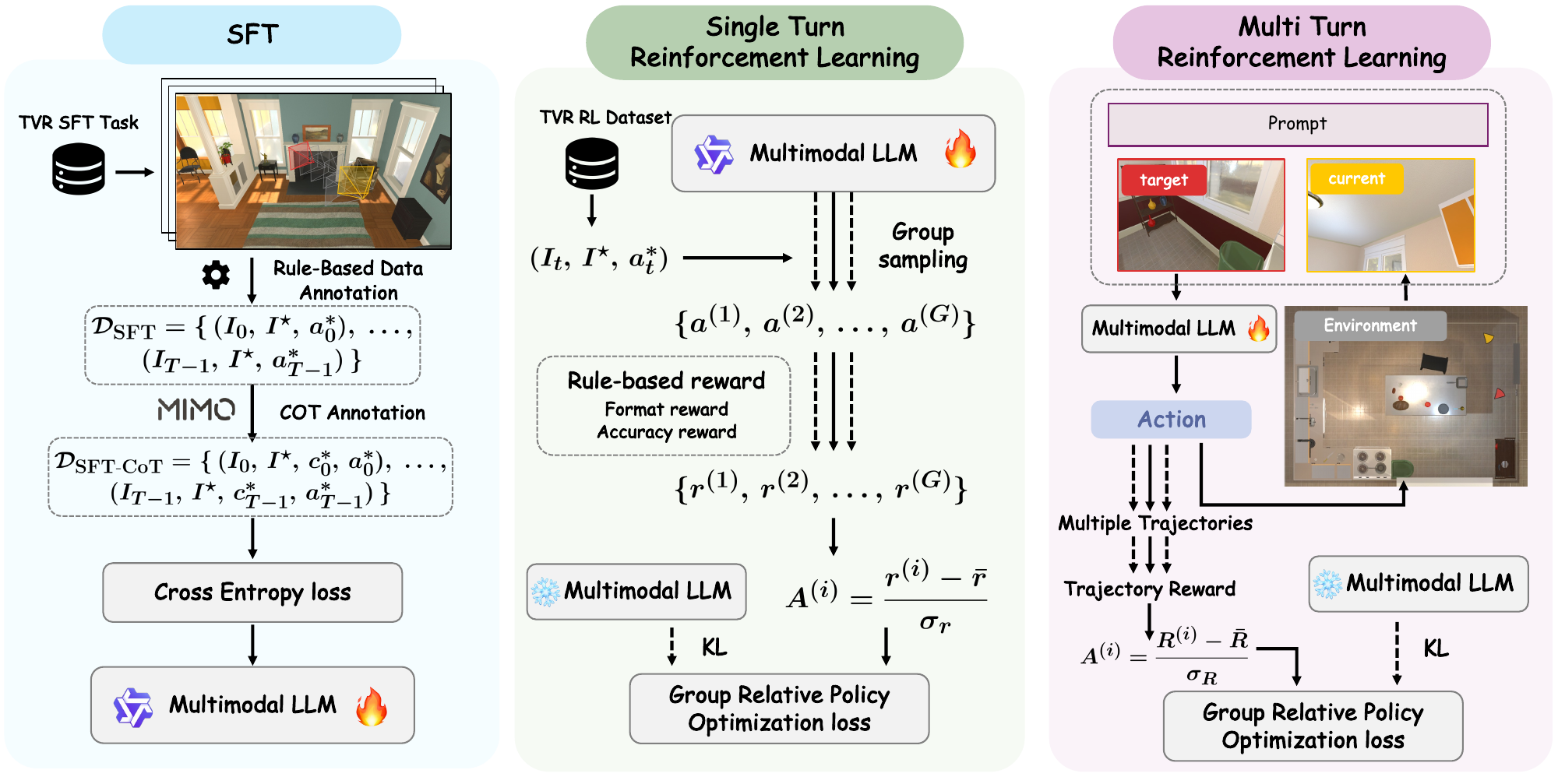}
\caption{\textbf{Post-training pipelines on TVRBench.} \emph{SFT}: supervised fine-tuning on rule-based expert trajectories (optionally with CoT). \emph{Single-turn RL}: GRPO on fixed $(I_t, I^\star, a^*_t)$ prompts. \emph{Multi-turn RL}: GRPO on on-policy rollouts in TVRBench with dense per-step plus terminal reward.}
\label{fig:posttraining_pipelines}
\end{figure*}

\begin{takeawaybox}
\itshape Across model scales, strategies, and memory formats, TVRBench reveals a consistent bottleneck: models struggle to map spatial discrepancies to reliable body translation, not visual matching.\end{takeawaybox}

\section{Can Post-Training Improve Active Viewpoint Control?}
\label{sec:post_training}
%

\begin{table*}[!t]
\centering
\small
\setlength{\tabcolsep}{4pt}
\setlength{\fboxsep}{2pt}
\caption{\textbf{Post-training results on TVRBench.} Success rate (\%) and diagnostics on the test split; Init names the SFT checkpoint each RL policy starts from; other columns follow Table~\ref{tab:fm_eval}; top-3 per column (excluding untrained baselines): \colorbox{rank1}{red}, \colorbox{rank2}{green}, \colorbox{rank3}{blue}.}
\label{tab:posttraining_main}
\begin{tabular}{lccccccccccc}
\toprule
\textbf{Method} & \textbf{Mem} & \textbf{CoT} & \textbf{Init} & \textbf{Overall}$\uparrow$ & \textbf{S-e}$\uparrow$ & \textbf{S-h}$\uparrow$ & \textbf{M-e}$\uparrow$ & \textbf{M-h}$\uparrow$ & \textbf{Steps} & \textbf{Stop} & \textbf{F-stop}$\downarrow$ \\
\midrule
Qwen3.5-9B                              &AO  & --  & --      & $2.8$            & $6.4$            & $4.8$            & $0.0$            & $0.0$            & $29.9$ & $27.0$ & $89.6$          \\
Qwen3.5-9B                              &VA  & --  & --      & $0.0$            & $0.0$            & $0.0$            & $0.0$            & $0.0$            & $34.3$ & $2.4$  & $100.0$         \\
\midrule
AO-CoT-SFT                              & AO  & yes & Qwen3.5-9B & $24.8$           & $40.8$           & $38.4$           & $8.8$            & $11.2$           & $30.5$ & $25.4$ & \cellcolor{rank2}$2.4$ \\
AO-SFT                                  & AO  & no  & Qwen3.5-9B & \cellcolor{rank3}$44.2$ & $61.6$           & \cellcolor{rank3}$60.8$ & \cellcolor{rank2}$32.0$ & \cellcolor{rank3}$22.4$ & $25.9$ & $48.0$ & \cellcolor{rank3}$7.9$ \\
VA-CoT-SFT                              & VA  & yes & Qwen3.5-9B & $35.6$           & \cellcolor{rank3}$68.0$ & $51.2$           & $12.8$           & $10.4$           & $28.9$ & $35.6$ & \cellcolor{rank1}$\mathbf{0.0}$ \\
\textbf{VA-SFT}                         & VA  & no  & Qwen3.5-9B & \cellcolor{rank2}$50.8$ & \cellcolor{rank1}$\mathbf{82.4}$ & \cellcolor{rank1}$\mathbf{68.8}$ & \cellcolor{rank3}$27.2$ & \cellcolor{rank2}$24.8$ & $25.3$ & $50.8$ & \cellcolor{rank1}$\mathbf{0.0}$ \\
\midrule
Single-turn GRPO                        & AO  & no  & AO-SFT  & $26.2$           & $48.8$           & $35.2$           & $12.0$           & $8.8$            & $28.2$ & $33.0$ & $20.6$ \\
\textbf{Multi-turn GRPO (ours)}         & VA  & no  & VA-SFT  & \cellcolor{rank1}$\mathbf{51.4}$ & \cellcolor{rank2}$81.6$ & \cellcolor{rank2}$64.0$ & \cellcolor{rank1}$\mathbf{34.4}$ & \cellcolor{rank1}$\mathbf{25.6}$ & $24.9$ & $51.4$ & \cellcolor{rank1}$\mathbf{0.0}$ \\
\bottomrule
\end{tabular}
\end{table*}

\paragraph{Setup.}
We use Qwen3.5-9B as the backbone for post-training.
For supervised fine-tuning (SFT), we vary the memory representation defined in Section~\ref{sec:tvrbench}, using either action-only (AO) or visual-action (VA) memory, and also vary whether the supervision contains intermediate Chain-of-Thought (CoT) rationales.
Training trajectories are produced by a rule-based expert in simulation.
For the CoT variants, MiMo-V2.5~\cite{mimo2026v25pro} provides the intermediate rationales through its API.
Appendix~\ref{app:sft_data} describes the annotation pipeline and dataset statistics.

We further apply Group Relative Policy Optimisation (GRPO) to the SFT checkpoints under two training setups (Figure~\ref{fig:posttraining_pipelines}).
Single-turn GRPO uses curated single-step prompts and an action-matching reward, while Multi-turn GRPO uses live TVRBench rollouts and an episode-level heuristic reward.
We use action-only memory for Single-turn GRPO and visual-action memory for Multi-turn GRPO, because visual-action memory is needed to retain the observation-action history required for trajectory-level optimisation.
Appendix~\ref{app:rl_setup} gives the RL data construction procedure, reward definitions, and hyperparameters.
All post-training checkpoints are evaluated on the TVRBench test split under the same step budgets as Section~\ref{sec:tvrbench}.
Table~\ref{tab:posttraining_main} reports per-category success rates and diagnostic metrics.

\paragraph{SFT Learns Action Mappings from Visual-Action History, Not CoT Rationales.}

Section~\ref{sec:fm_eval} identified a central bottleneck in TVR: 
models often fail to map spatial discrepancies to reliable embodied actions, especially body translation.
Supervised fine-tuning on expert trajectories substantially improves this discrepancy-to-action mapping across memory formats, 
with visual-action memory yielding the strongest results.

The best SFT setting, VA-SFT without CoT, reaches $50.8\%$ overall success on
TVRBench (Table~\ref{tab:posttraining_main}), far above both the untrained
Qwen3.5-9B baseline and the strongest closed-source baseline. Performance is
especially strong on single-room tasks (Single-easy $82.4\%$, Single-hard
$68.8\%$), while multi-room performance remains lower (Multi-easy $27.2\%$,
Multi-hard $24.8\%$), leaving the main room for further improvement.

The SFT ablations show two consistent trends: visual-action memory improves SFT performance, 
while our CoT rationales do not help. 
Without CoT, switching from action-only to visual-action memory raises overall success from $44.2\%$ to $50.8\%$; 
with CoT, the same switch raises success from $24.8\%$ to $35.6\%$. 
Conversely, adding CoT reduces success under both memory formats, 
from $44.2\%$ to $24.8\%$ with action-only memory and from $50.8\%$ to $35.6\%$ with visual-action memory. 
Stop calibration follows the same direction: both visual-action variants have F-stop $=0\%$, 
whereas action-only variants still make false Stop decisions in $2.4$--$7.9\%$ of Stop invocations. 
Thus, the SFT results identify visual-action memory as the more reliable ingredient for TVR, 
while CoT supervision is not beneficial under our current annotation scheme.

The degradation suggests that these rationales do not provide useful supervision
for this control policy, and may interfere with action learning under the current
annotation scheme. Whether CoT supervision tailored specifically to active
viewpoint control can help remains an open question.

\paragraph{Trajectory-level GRPO selectively improves multi-room exploration, whereas Single-turn GRPO regresses.}
Although the aggregate gain over VA-SFT is modest ($+0.6$\,pp), the split-level results are more informative. 
Multi-turn GRPO improves the long-distance multi-room splits, where SFT remains weakest: Multi-easy rises from $27.2$ to $34.4$ ($+7.2$\,pp), and Multi-hard from $24.8$ to $25.6$ ($+0.8$\,pp). 
The single-room splits remain close to the SFT checkpoint, with Single-easy changing from $82.4$ to $81.6$ and Single-hard from $68.8$ to $64.0$.
Thus, the benefit of Multi-turn GRPO is selective rather than uniform: it is most visible on the harder multi-room settings, while the stronger single-room performance is not substantially degraded.
The final model also keeps F-stop at $0\%$, suggesting that the multi-room gains do not come at the cost of worse stop calibration.

By contrast, casting the same RL data into single-step action-matching prompts consistently degrades the SFT policy.
Starting from AO-SFT, Single-turn GRPO reduces overall success by $18.0$\,pp ($44.2 \to 26.2$, Table~\ref{tab:posttraining_main}).
Starting from AO-CoT-SFT, it also reduces success by $9.8$ to $15.4$\,pp across KL coefficients $\beta \in \{0.01, 0.05\}$.
Stop calibration deteriorates as well: F-stop rises from $7.9\%$ at the AO-SFT initialization to $20.6\%$ after Single-turn GRPO.
Together, these results suggest that TVR-style active tasks benefit from RL only when the optimization objective matches their closed-loop, multi-step structure.
Per-step action matching is insufficient here and can degrade the supervised policy, whereas trajectory-level Multi-turn GRPO gives its clearest gains on the long-distance multi-room settings.

\begin{takeawayboxgreen}
\itshape
Most post-training gains on TVRBench come from visual-action SFT. Multi-turn GRPO adds a smaller, targeted refinement, with its clearest benefit on the long-distance multi-room splits.
\end{takeawayboxgreen}
\section{Conclusion}
We introduced \textbf{Target Viewpoint Reproduction (TVR)},
a closed-loop task for reproducing a target image
through embodied movement/reorientation,
and \textbf{TVRBench},
spanning scene scale and target-view visual richness.
TVRBench exposes a large model--human gap:
best closed/open-source models reach $12.0\%$/$7.8\%$ success
versus $93.0\%$ humans,
with failures mainly in mapping viewpoint discrepancies
to reliable body movement.
We further build a unified TVR post-training framework
covering expert-trajectory SFT, CoT-SFT,
Single-turn GRPO, and trajectory-level Multi-turn GRPO.
Visual-action SFT plus Multi-turn GRPO lifts a 9B model
from $2.8\%$ to $51.4\%$,
while CoT and Single-turn GRPO hurt closed-loop performance.
Together, TVR, TVRBench, and the post-training framework
provide a compact testbed for improving foundation models
that actively perceive and act in 3D.

\section*{Limitations}

TVRBench is built entirely in simulation (AI2-THOR and ProcTHOR-10k) with a discrete pose grid and an exact-pose success criterion.
These choices keep task difficulty controllable and the success signal unambiguous, but our results therefore characterize this setting rather than continuous, tolerance-based viewpoint control in the physical world.
Our post-training conclusions also rest on a single 9B open-source backbone, and we have not established how broadly they hold across model families, scales, and other active-perception tasks.

\section*{Ethical Considerations}

TVRBench builds on the AI2-THOR and ProcTHOR simulators and on MiMo-V2.5, GPT-4o, GPT-5, and Gemini-3.1-Pro (accessed via API), all used within their stated terms; human performance was collected from five volunteers.
We will release TVRBench, the trajectory pipeline, our post-training checkpoints, and supporting code under permissive open-source licenses.
TVR is evaluated entirely in indoor simulation; agents that actively control viewpoints could in principle enable intrusive uses, so any real-world deployment should be paired with domain-specific safety evaluation.



\bibliography{custom}

\appendix


\clearpage

\section{Appendix Overview}
\label{app:overview}

The appendix extends the main paper along five axes:

\begin{itemize}
\setlength\itemsep{2pt}
\item \textbf{Appendix~\ref{app:tvrbench}: TVRBench construction.} The $1{:}2{:}3$ scene split into SFT, evaluation, and RL pools, the per-category task generation procedure, the nine-action space, the four diagnostic metrics used throughout the paper, and the human evaluation protocol behind the human reference row.
\item \textbf{Appendix~\ref{app:sft_data}: SFT data pipeline.} Rule-based expert trajectories in simulation, Chain-of-Thought annotation with MiMo-V2.5, and the two memory formats---action-only and visual-action---that the SFT ablation crosses.
\item \textbf{Appendix~\ref{app:rl_setup}: Post-training configuration.} Hyperparameters and data construction for supervised fine-tuning, Single-turn GRPO, and Multi-turn GRPO, including the heuristic reward and action mask used in the multi-turn rollouts.
\item \textbf{Appendix~\ref{app:results}: Additional quantitative results.} A KL ablation for Single-turn GRPO, an RL-from-base bootstrap experiment, and a comparison between per-step matching accuracy and closed-loop episode success.
\item \textbf{Appendix~\ref{app:qualitative}: Qualitative examples.} Four representative trajectories on TVRBench---two failure modes of the untrained 9B (a rotation loop and a walking loop) and two successes of our VA-SFT + Multi-turn GRPO policy (one single-room iTHOR, one multi-room ProcTHOR).
\end{itemize}

\section{TVRBench Construction Details}
\label{app:tvrbench}

\subsection{Scene splits}
\label{app:tvrbench-scenes}
TVRBench uses 240 distinct indoor scenes split into post-training (SFT), evaluation, and reinforcement-learning (RL) pools at a $1{:}2{:}3$ ratio applied independently to each scene family to preserve the same family balance across pools.
The single-room half draws 120 scenes from AI2-THOR, with 30 each from its four scripted room categories---kitchen, living room, bedroom, and bathroom---yielding 20 SFT / 40 evaluation / 60 RL scenes.
The multi-room half draws 120 scenes uniformly at random from the training split of ProcTHOR-10k (each a procedurally generated 2--3 room layout with physical wall separation between rooms), partitioned under the same $1{:}2{:}3$ split.
No scene is shared across the three pools, ensuring that held-out evaluation tasks are drawn from environments unseen during both SFT and RL training, so reported results reflect genuine generalisation rather than scene memorisation.

\subsection{Task generation}
\label{app:tvrbench-tasks}
Each TVRBench task is a (start, target) pose pair sampled within a single scene and characterised by two independent dimensions: (i) the shortest-path length between start and target on the agent's discrete pose graph---the minimum number of unit actions a rule-based expert needs to navigate from one to the other, which proxies the spatial extent of the navigation required---and (ii) the segment count $\mathrm{seg}$ at the target viewpoint, computed as the number of visible objects excluding structural geometry, the agent itself, and meshes flagged by an internal exclusion list, a value that proxies the visual richness of the target viewpoint.
Crossing the two dimensions yields the four task categories used throughout the paper: single-room easy ($\mathrm{seg} \geq 9$, path length $2$--$8$) and single-room hard ($\mathrm{seg} \in [3, 6]$, path length $2$--$8$), both drawn from AI2-THOR scenes; multi-room easy ($\mathrm{seg} \geq 9$, path length $10$--$20$) and multi-room hard ($\mathrm{seg} \in [3, 6]$, path length $10$--$20$), both drawn from ProcTHOR-10k scenes.
The intermediate band $\mathrm{seg} \in [7, 8]$ is held out as a gap to keep the easy and hard tiers clearly separated.
The total number of generated tasks is 1{,}600 for SFT (40 per scene over the 40 SFT scenes), 500 for evaluation (125 per category), and 4{,}800 for RL (40 per scene over the 120 RL scenes).

\subsection{Action space}
\label{app:tvrbench-actions}
At each step the agent selects one of nine discrete actions on AI2-THOR's discrete pose grid: four agent-frame translations (\texttt{MoveAhead}, \texttt{MoveBack}, \texttt{MoveLeft}, \texttt{MoveRight}) by $0.25$\,m, two body rotations (\texttt{RotateLeft}, \texttt{RotateRight}) by $\pm 45^\circ$ about the vertical axis (refined from the simulator's $90^\circ$ default for finer viewpoint control), two head pitches (\texttt{LookUp}, \texttt{LookDown}) by $\pm 30^\circ$ within the simulator's $[-30^\circ, +30^\circ]$ horizon range, and a single termination action (\texttt{Stop}).
The simulator rejects any action that would result in a collision with scene geometry; in such cases the pose is unchanged but the step still counts against the per-task budget, which discourages blind movement into obstacles.
Episodes terminate either when the agent issues Stop (success requires that this happens at a target-matching pose) or when the step budget is exhausted, the latter counted as a failure.

\subsection{Diagnostic metrics}
\label{app:tvrbench-metrics}
Let $\mathcal{E} = \{e_i\}_{i=1}^{N}$ be a set of $N$ evaluation episodes, with per-episode quantities $S_i \in \{0,1\}$ (success per the criterion in Section~\ref{app:tvrbench-actions}), $\mathrm{stop}_i \in \{0,1\}$ (whether Stop was issued), the pose-match indicator $m_i \in \{0,1\}$ (whether $s_T = s^{\star}$, so that $S_i = \mathrm{stop}_i\, m_i$), $T_i$ (number of actions taken), and the final-step pose errors $|\Delta p|_i$, $|\Delta \theta|_i$, $|\Delta \varphi|_i$ defined in Section~\ref{sec:tvrbench}.
We report:
\begin{align*}
\mathrm{SR}              &= \frac{1}{N}\sum_{i} S_i,                                       \\
\mathrm{Steps}            &= \frac{1}{N}\sum_{i} T_i,                                       \\
\mathrm{F\text{-}stop}    &= \frac{\sum_{i} \mathrm{stop}_i\,(1 - m_i)}{\sum_{i} \mathrm{stop}_i},     \\
|\overline{\Delta x}|     &= \frac{1}{N}\sum_{i} |\Delta x|_i, \quad x \in \{p,\theta,\varphi\}.
\end{align*}
The per-category rates S-e, S-h, M-e, M-h are obtained by restricting the sum to episodes in the respective category (each has $N=125$ in the evaluation split), which isolates performance on each difficulty tier.
$\mathrm{F\text{-}stop}$, the false-stop rate, is conditioned on the episodes that stop: it measures how often Stop is issued at a non-target pose among them, so a model that rarely stops can still have a high $\mathrm{F\text{-}stop}$ if those few stops are wrong, and a low value should be read together with the Stop rate.

\subsection{Human evaluation protocol}
\label{app:tvrbench-human}
To establish a human reference point, five volunteers each completed a balanced 100-task subset of the evaluation split, with 25 tasks drawn from each of the four categories so that scene scale and visual richness are equally represented.
Participants drove the agent through a single-user web interface that displays the current first-person observation and the target image side by side, and issued the same nine discrete actions available to the models through a fixed keyboard mapping: \texttt{W}/\texttt{S}/\texttt{A}/\texttt{D} for the four translations, \texttt{Q}/\texttt{E} for body rotation, \texttt{R}/\texttt{F} for head pitch, and the space bar for \texttt{Stop}.
They were instructed to reproduce the target viewpoint as closely as possible and then press Stop to declare completion.
Every run used exactly the same image resolution, action space, per-task step budget ($30$ actions for single-room iTHOR tasks and $40$ for multi-room ProcTHOR tasks), and pose-matching success criterion ($|\Delta p| \leq 0.01$\,m, $|\Delta\theta| \leq 1^\circ$, $|\Delta\varphi| \leq 1^\circ$) as the model evaluation, so the human and model rows are directly comparable.
Participation was voluntary and unpaid, and participants were informed that their anonymous task performance would be used solely as the human reference reported in this paper.
The task involves only navigating an indoor simulator and poses no foreseeable risk to participants, so no risk disclaimers were required.

\section{SFT Data Pipeline}
\label{app:sft_data}

\subsection{Rule-based trajectory generation}
\label{app:sft-traj}
Expert trajectories for the SFT pool are produced offline by an oracle planner with \emph{privileged access} to simulator-internal state: the agent's exact pose, the precomputed reachable-position graph of each scene, and the target pose.
This information is unavailable to any of the learned models we evaluate.
For each task $(s_0, s^\star)$ the planner emits a three-phase action sequence:
\begin{enumerate}
\item \textbf{View alignment.} Rotate the body and adjust head pitch from $(\theta_0, \varphi_0)$ to $(\theta^\star, \varphi^\star)$ using the minimum number of \texttt{RotateLeft}/\texttt{Right} and \texttt{LookUp}/\texttt{Down} actions.
\item \textbf{Navigation.} Run Dijkstra's shortest-path algorithm on the discrete state space $(x, z, \theta)$ from $(p_0, \theta^\star)$ to $(p^\star, \theta^\star)$, where each of the six body-motion actions (\texttt{MoveAhead}/\texttt{Back}/\texttt{Left}/\texttt{Right}, \texttt{RotateLeft}/\texttt{Right}) is a unit-cost edge. The agent is permitted to rotate away from $\theta^\star$ during navigation but must end at $\theta^\star$, which avoids inefficient zig-zag motion toward off-axis targets.
\item \textbf{Termination.} Issue \texttt{Stop}.
\end{enumerate}
The planner is deterministic and produces exactly one minimum-action-count trajectory per task, for a total of 1{,}600 SFT trajectories whose lengths equal the shortest-path lengths used to define task difficulty (Section~\ref{app:tvrbench-tasks}), so every demonstration is action-optimal by construction.
The trajectories serve solely as the supervision target for SFT; the learned policy operates strictly from first-person observations and never receives the privileged state used by the planner.

\subsection{CoT annotation with MiMo-V2.5}
\label{app:sft-cot}

\begin{figure*}[!tb]
\centering
\begin{promptbox}[CoT Annotation Prompt]
The ground-truth action is: \texttt{\{a*\_t\}}.\\
This action is CORRECT. Your task is to JUSTIFY it---not propose a different one. Even if you would have picked another action, your rationale MUST logically support \texttt{`\{a*\_t\}`}.\\[6pt]

Write a SHORT observation-grounded rationale (1--3 sentences):
\begin{itemize}\itemsep0pt
\item briefly note 1--2 key objects/landmarks visible in the CURRENT view,
\item compare with the TARGET (what's misaligned: heading, distance, or position),
\item conclude why \texttt{`\{a*\_t\}`} reduces that gap (do NOT name any other action verb).
\end{itemize}
Only mention objects you actually see (do not invent or guess).\\[6pt]

Output STRICT JSON: \texttt{\{"reasoning": "<your rationale here>"\}}---no markdown, no prefix like \texttt{Action:} or \texttt{Rationale:}; the value must be a plain prose string.
\end{promptbox}
\caption{Instructions appended to the per-step SFT user message at step $t$ (containing the current observation $I_t$, target image $I^\star$, and action history) when querying MiMo-V2.5 for a chain-of-thought rationale. \texttt{\{a*\_t\}} is the expert action returned by the rule-based planner.}
\label{fig:cot-prompt}
\end{figure*}

For the CoT variants (AO-CoT-SFT and VA-CoT-SFT), we augment the rule-based trajectories with intermediate chain-of-thought rationales.
For each (current observation $I_t$, target image $I^\star$, expert action $a^*_t$) triple produced by the planner, we prompt the MiMo-V2.5 model (accessed via API) to write a short, observation-grounded justification of why $a^*_t$ is correct, keeping every rationale consistent with the optimal action label.
The prompt (Figure~\ref{fig:cot-prompt}) instructs the model to (i) reference 1--2 visible landmarks in the current view, (ii) compare them against the target and identify the misalignment dimension (heading, distance, or position), and (iii) explain how the given action reduces that gap, without naming any alternative action.
The 1--3 sentence cap is deliberately tight: SFT trajectories preserve the full multi-turn history of observations and reasoning across up to $30$--$40$ steps, so any per-step rationale length is multiplied by the trajectory length when accumulated in context.
We accept the returned rationale only if it parses as the requested JSON object, discarding any malformed response.

\subsection{Two memory formats}
\label{app:sft-format}

\begin{figure*}[!tb]
\centering
\begin{promptbox}[Action-Only Memory Sample (per-step)]
\textbf{SYSTEM}\\
You are a navigation agent in an indoor environment. Your task is to navigate and adjust your viewpoint to PRECISELY match a target image. You must match the exact position, orientation, and camera angle---the goal is for your observation to look identical to the target.\\[2pt]
Available actions:\\
\hspace*{1em}\texttt{- MoveAhead: Move forward 0.25m}\\
\hspace*{1em}\texttt{- MoveBack: Move backward 0.25m}\\
\hspace*{1em}\texttt{- MoveLeft: Move left 0.25m}\\
\hspace*{1em}\texttt{- MoveRight: Move right 0.25m}\\
\hspace*{1em}\texttt{- RotateRight: Rotate clockwise 45\textdegree}\\
\hspace*{1em}\texttt{- RotateLeft: Rotate counter-clockwise 45\textdegree}\\
\hspace*{1em}\texttt{- LookUp: Tilt camera up 30\textdegree}\\
\hspace*{1em}\texttt{- LookDown: Tilt camera down 30\textdegree}\\
\hspace*{1em}\texttt{- Stop: Declare that you have reached the target viewpoint}\\[2pt]
You will receive: (1) your current observation; (2) your recent action history (if available); (3) the target viewpoint you need to match. Use your action history to avoid repeating ineffective actions. You MUST respond in exactly this format: ``\texttt{Action: <action name>}''.\\[6pt]

\textbf{USER}\\
Your CURRENT observation:\quad [image: $I_t$]\\
Your recent actions (5 steps):\\
\hspace*{1em}- Step 5: MoveAhead\\
\hspace*{1em}- Step 6: MoveAhead\\
\hspace*{1em}- Step 7: RotateRight\\
\hspace*{1em}- Step 8: MoveAhead\\
\hspace*{1em}- Step 9: MoveAhead\\
TARGET viewpoint you must match:\quad [image: $I^\star$]\\
Valid actions at this step: MoveAhead, MoveBack, RotateLeft, RotateRight, LookUp, LookDown, Stop.\\[6pt]

\textbf{GPT}\\
<think> The target view shows the kitchen counter centred and close, while my current view has it to the right of centre. Rotating right will align my heading with the target. </think>\\
Action: RotateRight.
\end{promptbox}
\caption{SFT sample format under \textbf{action-only} memory. Each trajectory step becomes one independent single-turn sample with a textual recent-action history; no past observations remain in context. The \texttt{<think>\ldots</think>} prefix appears only in CoT variants. ``\texttt{Valid actions at this step:}'' lists the actions the simulator allows at the current pose. The $1{,}600$ SFT trajectories expand to $\approx 20{,}700$ such per-step samples.}
\label{fig:sample-format-ao}
\end{figure*}

\begin{figure*}[!tb]
\centering
\begin{promptbox}[Visual-Action Memory Sample (multi-turn, single trajectory)]
\textbf{SYSTEM}\quad (identical to the prompt above)\\[6pt]

\textbf{USER (turn 0)}\\
Your CURRENT observation:\quad [image: $I_0$]\\
TARGET viewpoint you must match:\quad [image: $I^\star$]\\
Valid actions at this step: MoveAhead, RotateLeft, RotateRight, LookUp, LookDown, Stop.\\
\textbf{GPT (turn 0)}\\
<think> The target view shows a kitchen counter on the left; my current view shows it diagonally ahead. I should turn left first to align my heading. </think>\\
Action: RotateLeft.\\[6pt]

\textbf{USER (turn 1)}\\
Your CURRENT observation:\quad [image: $I_1$]\\
TARGET viewpoint you must match:\quad [image: $I^\star$]\\
Valid actions at this step: MoveAhead, MoveBack, RotateLeft, RotateRight, LookUp, LookDown, Stop.\\
\textbf{GPT (turn 1)}\\
<think> Counter now sits straight ahead but the target view is closer. I need to move forward. </think>\\
Action: MoveAhead.\\[6pt]

\textbf{USER (turn 2)}\\
Your CURRENT observation:\quad [image: $I_2$]\\
TARGET viewpoint you must match:\quad [image: $I^\star$]\\
Valid actions at this step: MoveAhead, MoveBack, RotateLeft, RotateRight, LookUp, LookDown, Stop.\\
\textbf{GPT (turn 2)}\\
<think> Closer, but I am still further than the target view; one more step ahead. </think>\\
Action: MoveAhead.\\[6pt]

\textbf{USER (turn 3)}\\
Your CURRENT observation:\quad [image: $I_3$]\\
TARGET viewpoint you must match:\quad [image: $I^\star$]\\
Valid actions at this step: MoveBack, RotateLeft, RotateRight, LookUp, LookDown, Stop.\\
\textbf{GPT (turn 3)}\\
<think> The view now matches the target in both heading and distance. Terminate. </think>\\
Action: Stop.
\end{promptbox}
\caption{SFT sample format under \textbf{visual-action} memory. The entire trajectory is packed into a single multi-turn sample; all past observations remain in context at every step (the SYSTEM prompt is identical to Figure~\ref{fig:sample-format-ao}). The \texttt{<think>\ldots</think>} prefix appears only in CoT variants. The trajectory shown is a 4-step example; in TVRBench, trajectories range from a few up to roughly 30--40 steps. This yields exactly $1{,}600$ multi-turn samples.}
\label{fig:sample-format-va}
\end{figure*}

The two memory representations produce structurally different SFT samples.
Under \textbf{action-only} memory each trajectory step becomes an independent single-turn sample whose user message contains the current observation $I_t$, the target image $I^\star$, and a short action-history text; under \textbf{visual-action} memory the entire trajectory is packed into a single multi-turn sample whose turns accumulate end to end, exposing every past observation in context at every step, so action-only keeps sequences short while visual-action preserves the full visual memory.
For the CoT variants, each model response is optionally prefixed by a chain-of-thought rationale wrapped in \texttt{<think>\ldots</think>} tags.
Figures~\ref{fig:sample-format-ao} and~\ref{fig:sample-format-va} give concrete schematics of each format.

\section{Post-Training Configuration}
\label{app:rl_setup}

\paragraph{Compute budget.}
A single supervised fine-tuning run uses $4$ NVIDIA H100 GPUs for roughly $6$ hours; Multi-turn (online) GRPO uses $8$ NVIDIA H200 GPUs for roughly $10$ hours; and Single-turn (offline) GRPO uses $8$ H200 GPUs for roughly $4$ hours.

\subsection{Supervised fine-tuning}
\label{app:pt-sft}
All four SFT variants fine-tune Qwen3.5-9B with full-parameter updates and a frozen vision encoder that preserves its pretrained visual representations.
We use AdamW with bf16 precision, learning rate $1\times10^{-5}$ under a cosine schedule with $10\%$ linear warmup, image resolution capped at $262\,144$ pixels, per-device batch size $1$ with gradient accumulation $8$ across $4$ GPUs (effective batch $32$), and DeepSpeed ZeRO-2 with gradient checkpointing.
Training runs for $3$ epochs on the AO variants (AO-SFT, AO-CoT-SFT) and $5$ epochs on the VA variants (VA-SFT, VA-CoT-SFT), whose far smaller sample count warrants the additional passes.

\subsection{Single-turn GRPO}
\label{app:pt-single}
Single-turn GRPO optimises a per-step action policy on a parquet dataset of $(I_t, I^\star, a^*_t)$ prompts, built by flattening the SFT trajectories into independent (state, expert-action) tuples, so each action is trained in isolation from its trajectory context.
The policy is initialised from the AO-SFT checkpoint, and we inherit the GRPO implementation from \texttt{verl}~\cite{shao2024deepseekmath}.

\paragraph{GRPO objective.}
For each prompt $q$, we sample a group of $G$ responses $\{o_i\}_{i=1}^{G}$ from the current rollout policy $\pi_{\mathrm{old}}$ and score each with a scalar reward $r_i$.
The group-relative advantage centres $r_i$ on the group mean and normalises by the group standard deviation,
\begin{gather*}
A_i = \frac{r_i - \bar{r}}{\sigma_r}, \\
\bar{r} = \tfrac{1}{G}\textstyle\sum_j r_j, \quad \sigma_r = \mathrm{std}(\{r_j\}),
\end{gather*}
and is broadcast to every token of $o_i$.
The GRPO objective adopts the PPO-style clipped surrogate over this advantage, together with a KL anchor against the SFT reference $\pi_{\mathrm{ref}}$,
\begin{gather*}
\mathcal{J}(\theta) = \mathbb{E}_{i,t}\!\left[\min\!\big(\rho_t A_i,\ \mathrm{clip}(\rho_t, 1{-}\epsilon, 1{+}\epsilon)\, A_i\big)\right] \\
- \beta\, D_{\mathrm{KL}}\!\left[\pi_\theta \,\|\, \pi_{\mathrm{ref}}\right],
\end{gather*}
where $\rho_t = \pi_\theta(o_{i,t}\mid q, o_{i,<t}) / \pi_{\mathrm{old}}(o_{i,t}\mid q, o_{i,<t})$ and $D_{\mathrm{KL}}$ is estimated with the unbiased low-variance K3 form.

\paragraph{Reward.}
The per-response reward $r_i$ is a gated combination of format validity and action correctness,
\[
r_i = \mathbf{1}\!\left[\mathrm{format}(o_i)\right] \cdot \big( 0.1 + 0.9\cdot \mathbf{1}\!\left[a(o_i) = a^*_t\right] \big),
\]
so a response that drops the required ``\texttt{Action:~<name>}'' format receives $0$, a correctly-formatted but wrong action receives $0.1$, and a correctly-formatted matching action receives $1.0$; the floor still rewards valid formatting even when the chosen action is wrong.

\paragraph{Hyperparameters.}
Group size $G=8$ rollouts at temperature $0.9$ and top-$p$ $0.95$. AdamW with learning rate $1\times10^{-6}$, gradient clip $1.0$, no entropy bonus. GRPO clip threshold $\epsilon=0.2$ (\texttt{verl} default); the KL coefficient is $\beta\in\{0.01,\ 0.05\}$, with $\beta=0.01$ reported as the default row in Table~\ref{tab:posttraining_main} and $\beta=0.05$ included in the KL ablation (Appendix~\ref{app:results-kl}).

\subsection{Multi-turn GRPO}
\label{app:pt-multi}
Multi-turn GRPO optimises an episode-level policy by rolling out trajectories in the live TVRBench simulator, learning on-policy from closed-loop interaction.
For each task (a $(\mathrm{start},\mathrm{target})$ pose pair drawn from a dedicated $4{,}800$-task RL split), the policy is rolled out $G$ times against the simulator; each rollout produces a trajectory
\[
\tau^{(i)} = (\mathrm{obs}_0,\, a_1, r_1, \,\ldots,\, a_{T_i}, r_{T_i}, \mathrm{obs}_{T_i}),
\]
with per-step rewards $r_t$ given below.
We initialise from the VA-SFT checkpoint, inherit the GRPO core from \texttt{verl}~\cite{shao2024deepseekmath}, and use a custom agent loop that interleaves model-generated actions with simulator observations.

\paragraph{Per-step reward.}
The reward at step $t$ decomposes additively into four components,
\[
r_t = -c_{\mathrm{step}} + r_{\mathrm{fmt}}^{(t)} + r_{\mathrm{prog}}^{(t)} + r_{\mathrm{term}}^{(t)},
\]
with: (i) a constant \emph{step penalty} $c_{\mathrm{step}} = 0.01$ to encourage efficiency; (ii) a \emph{format} term $r_{\mathrm{fmt}}^{(t)} = +0.005$ if the model output parses to a valid action, $-0.01$ otherwise; (iii) an asymmetric \emph{progress} term that only rewards strict improvements in the running minimum pose distance,
\[
r_{\mathrm{prog}}^{(t)} = \max\!\big\{0,\ d_{\min}^{(t-1)} - d_t\big\},
\]
where $d_{\min}^{(t-1)} = \min_{s \leq t-1} d_s$ tracks the best distance seen so far, so backtracking toward already-visited poses earns no reward; and (iv) a \emph{terminal} term $r_{\mathrm{term}}^{(t)} = +1.0$ when the agent issues Stop at the target pose, $-0.5$ when it issues Stop at a non-target pose, and $0$ otherwise, so a premature or mistaken Stop is actively penalised rather than merely left unrewarded.
The pose distance is a weighted geodesic
\[
d_t = \|p_t - p^\star\|_2 + 0.25\,n^{\mathrm{rot}}_t + 0.25\,n^{\mathrm{hor}}_t,
\]
where $n^{\mathrm{rot}}_t = \min(|\Delta\theta_t|, 360^\circ\!-\!|\Delta\theta_t|)/45^\circ$ and $n^{\mathrm{hor}}_t = |\Delta\varphi_t|/30^\circ$ are the integer numbers of rotation and head-pitch actions needed to align with the target, weighted so that one such action contributes the same as one $0.25\,\mathrm{m}$ translation step.

\paragraph{Trajectory-level advantage.}
The scalar reward attributed to each rollout is the sum of its per-step rewards,
\[
R^{(i)} = \sum_{t=1}^{T_i} r_t^{(i)},
\]
and the group-relative advantage is computed at the trajectory level and broadcast to every assistant token of $\tau^{(i)}$,
\begin{gather*}
A^{(i)} = \frac{R^{(i)} - \bar{R}}{\sigma_R}, \\
\bar{R} = \tfrac{1}{G}\textstyle\sum_j R^{(j)}, \quad \sigma_R = \mathrm{std}(\{R^{(j)}\}).
\end{gather*}

\paragraph{Token-masked objective.}
Because each trajectory interleaves environment observations with model-generated actions, only assistant tokens carry gradients, so the policy is never trained to predict the simulator's observations.
We mask the GRPO loss accordingly,
\begin{gather*}
\mathcal{J}(\theta) = \mathbb{E}\!\left[m_t \cdot \min\!\big(\rho_t A^{(i)},\right. \\
\left.\mathrm{clip}(\rho_t, 1{-}\epsilon, 1{+}\epsilon)\, A^{(i)}\big)\right] \\
- \beta\, D_{\mathrm{KL}}(\pi_\theta \,\|\, \pi_{\mathrm{ref}}),
\end{gather*}
where $m_t = \mathbf{1}[\text{token}_t \in \text{assistant}]$ and $\pi_{\mathrm{ref}}$ is the VA-SFT initialisation.

\paragraph{Hyperparameters.}
Group size $G=8$ trajectories per task, maximum trajectory length $T_{\max}=30$ turns, with $8$ parallel environment instances per rollout worker.
AdamW with learning rate $1\times10^{-7}$---an order of magnitude smaller than the Single-turn case to preserve the stronger VA-SFT initialisation---and gradient clip $1.0$.
GRPO clip $\epsilon=0.2$ and KL coefficient $\beta=0.01$, both inherited from the Single-turn configuration (Appendix~\ref{app:pt-single}).

\section{Additional Quantitative Results}
\label{app:results}

\subsection{KL ablation for Single-turn GRPO}
\label{app:results-kl}
We expand the claim from Section~\ref{sec:post_training} that ``even the most KL-conservative setting still regresses below its SFT initialisation'' with the per-category breakdown in Table~\ref{tab:kl-ablation}.
Starting from AO-CoT-SFT, Single-turn GRPO drops by $-9.8$\,pp at $\beta=0.05$ and by $-15.4$\,pp at the more permissive $\beta=0.01$.
Both settings also degrade the stop calibration: F-stop rises from $2.4\%$ at the SFT init to $10.9\%$ ($\beta=0.05$) and $23.5\%$ ($\beta=0.01$), so both success rate and stop calibration worsen monotonically as the KL leash is loosened.

\begin{table}[h]
\centering
\footnotesize
\setlength{\tabcolsep}{3pt}
\caption{KL ablation for Single-turn GRPO initialised from AO-CoT-SFT. Both $\beta$ settings regress below the SFT init and worsen F-stop calibration.}
\label{tab:kl-ablation}
\begin{tabular*}{\columnwidth}{@{\extracolsep{\fill}}lccccc@{}}
\toprule
\textbf{Config} & \textbf{Overall}$\uparrow$ & \textbf{S-e}$\uparrow$ & \textbf{S-h}$\uparrow$ & \textbf{M-e}$\uparrow$ & \textbf{M-h}$\uparrow$ \\
\midrule
AO-CoT-SFT (init)   & $24.8$  & $40.8$ & $38.4$ & $8.8$ & $11.2$ \\
\quad $+\beta=0.05$ & $15.0$  & $32.8$ & $19.2$ & $4.8$ & $3.2$  \\
\quad $+\beta=0.01$ & $9.4$   & $20.8$ & $12.8$ & $1.6$ & $2.4$  \\
\bottomrule
\end{tabular*}
\end{table}

\subsection{RL from a base initialisation}
\label{app:results-bootstrap}
We also evaluated both GRPO variants without any SFT warm-up, starting directly from the untrained Qwen3.5-9B (Table~\ref{tab:rl-bootstrap}).
Single-turn GRPO improves the AO baseline only marginally ($2.8 \to 3.6$, $+0.8$\,pp); Multi-turn GRPO, by contrast, lifts the VA baseline from $0\%$ to $\mathbf{26.2\%}$ overall and achieves perfect stop calibration (F-stop $=0\%$).
Trajectory-level on-policy RL alone produces a workable policy from scratch, whereas per-step RL does not, because the shaped progress reward supplies a learning signal even to a near-random initial policy.

\begin{table}[h]
\centering
\footnotesize
\setlength{\tabcolsep}{2pt}
\caption{GRPO from a base (no SFT) initialisation. Multi-turn GRPO bootstraps a workable policy from the untrained Qwen3.5-9B (VA); Single-turn GRPO does not.}
\label{tab:rl-bootstrap}
\begin{tabular*}{\columnwidth}{@{\extracolsep{\fill}}lccccc@{}}
\toprule
\textbf{Config} & \textbf{Overall}$\uparrow$ & \textbf{S-e}$\uparrow$ & \textbf{S-h}$\uparrow$ & \textbf{M-e}$\uparrow$ & \textbf{M-h}$\uparrow$ \\
\midrule
Qwen3.5-9B base (AO)           & $2.8$            & $6.4$  & $4.8$  & $0.0$  & $0.0$ \\
\quad $+$ Single-turn GRPO     & $3.6$            & $11.2$ & $3.2$  & $0.0$  & $0.0$ \\
Qwen3.5-9B base (VA)           & $0.0$            & $0.0$  & $0.0$  & $0.0$  & $0.0$ \\
\quad $+$ Multi-turn GRPO      & $\mathbf{26.2}$  & $52.8$ & $31.2$ & $15.2$ & $5.6$ \\
\bottomrule
\end{tabular*}
\end{table}

\subsection{Per-step versus closed-loop accuracy}
\label{app:results-diag}
We check whether the Single-turn GRPO closed-loop regression is masked by per-step gains.
Replaying the validation split of the per-step prompt dataset ($500$ prompts $\times$ $8$ rollouts) through the AO-CoT-SFT $+$ Single-turn GRPO checkpoint ($\beta=0.01$, step $100$) yields a per-step action-matching accuracy of $72.1\%$, with format validity $99.98\%$ and an average per-step reward of $0.749$; a parallel run at $\beta=0.001$ produces a comparable per-step accuracy of $0.78$.
The same $\beta=0.01$ checkpoint, however, reaches only $9.4\%$ on the closed-loop benchmark (Table~\ref{tab:posttraining_main}), a gap of over $60$ points between per-step matching and episode success.
The gap reflects compounding error: small per-step inaccuracies accumulate over the roughly $30$ decisions per episode, and the policy never learns recovery, as it is trained only on expert-conditioned states, not the off-expert states it visits at test time.
A per-step matching objective therefore does not translate into end-to-end trajectory success.

\section{Qualitative Examples}
\label{app:qualitative}

We complement the aggregate numbers in Sections~\ref{sec:fm_eval}--\ref{sec:post_training} with four end-to-end TVRBench traces: two failures of the untrained Qwen3.5-9B (one rotating in place, one walking in a short loop) and two successes of our VA-SFT + Multi-turn GRPO policy (single-room iTHOR and multi-room ProcTHOR).
Each trace shows an orthographic floor plan with start (yellow), target (red), and final pose (blue), the full path, and first-person frames sampled along the trajectory.

\begin{figure*}[!t]
\centering
\includegraphics[width=0.97\textwidth]{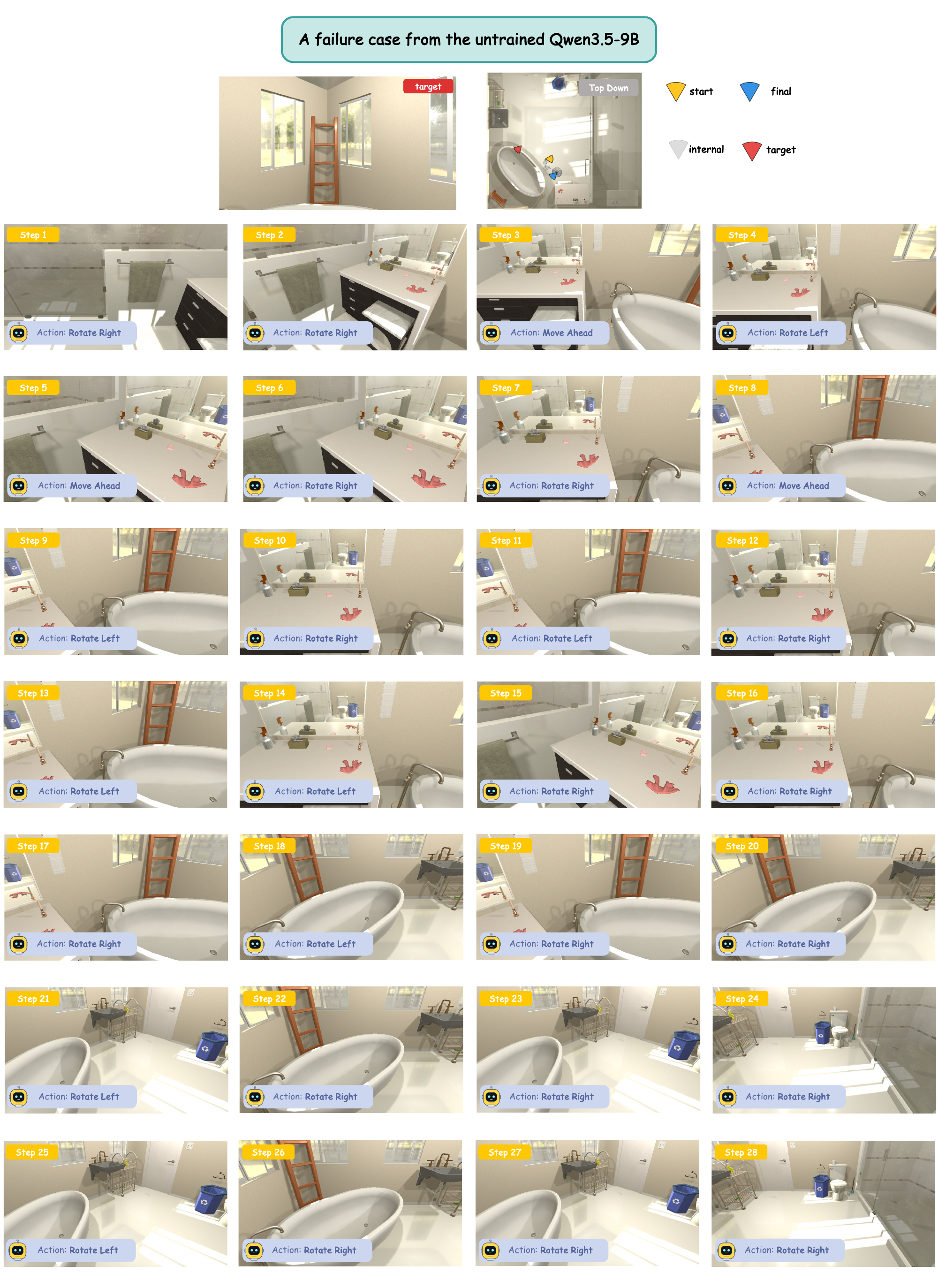}
\caption{\textbf{A failure case from the untrained Qwen3.5-9B.} With action-only memory, the agent advances twice in its first four steps and then issues 35 consecutive Rotate actions at the same position until the 40-step budget runs out. The action history alone cannot tell the policy it has already tried---and rejected---each yaw, so the same micro-decision repeats indefinitely. For space, the panels show only the first 28 of 40 steps, which continue the same in-place rotation.}
\label{fig:qual-fail-case1}
\end{figure*}

\begin{figure*}[!t]
\centering
\includegraphics[width=0.97\textwidth]{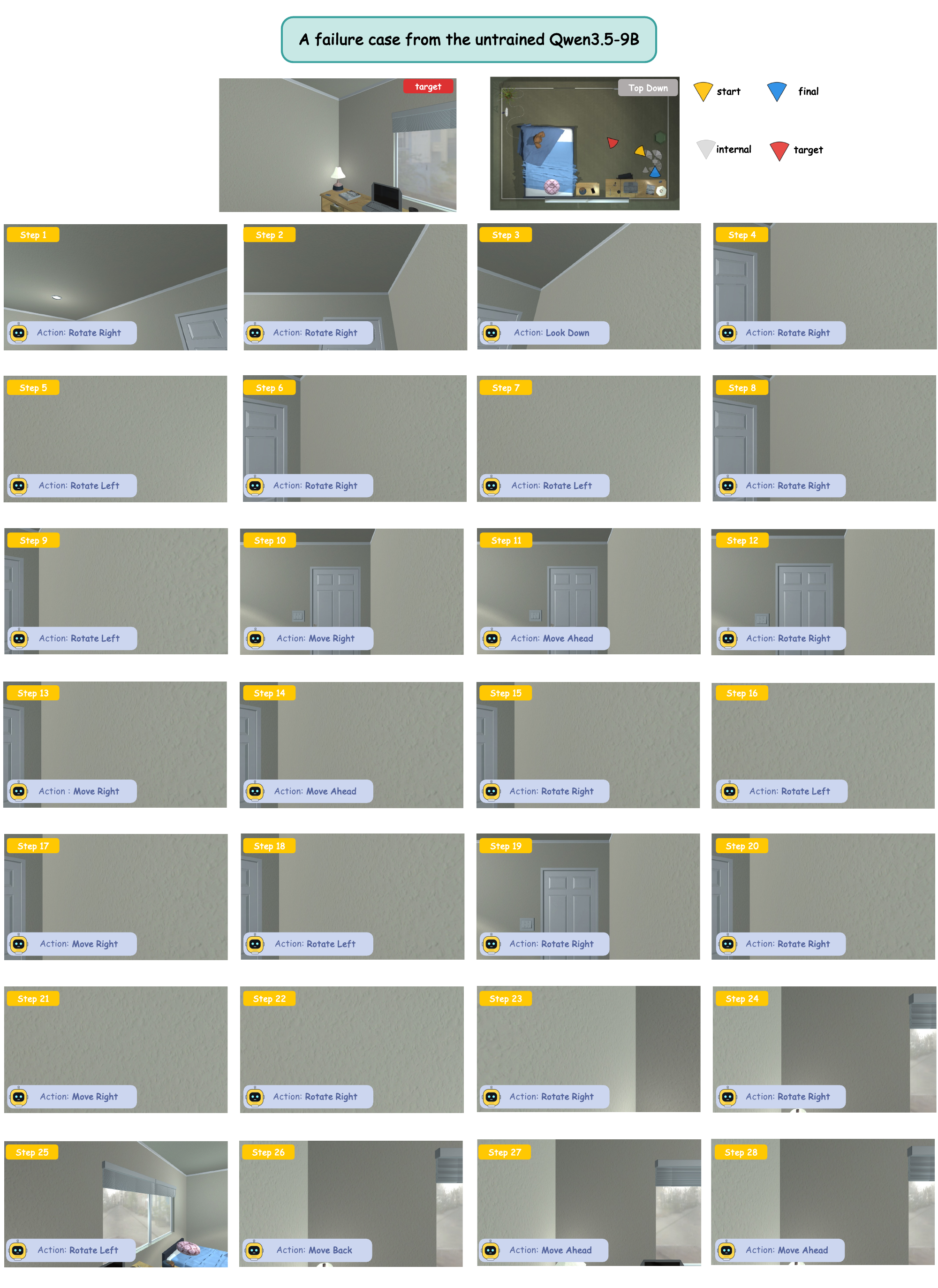}
\caption{\textbf{A failure case from the untrained Qwen3.5-9B.} On a single-room iTHOR scene the agent shuttles between a handful of cells---issuing 12 Move actions among only four distinct positions---without ever closing the gap to the target. The action history alone cannot register that these cells have already been visited, so the same short walking loop repeats. For space, the panels show only the first 28 of 30 steps.}
\label{fig:qual-fail-case2}
\end{figure*}

\begin{figure*}[!t]
\centering
\includegraphics[width=0.97\textwidth]{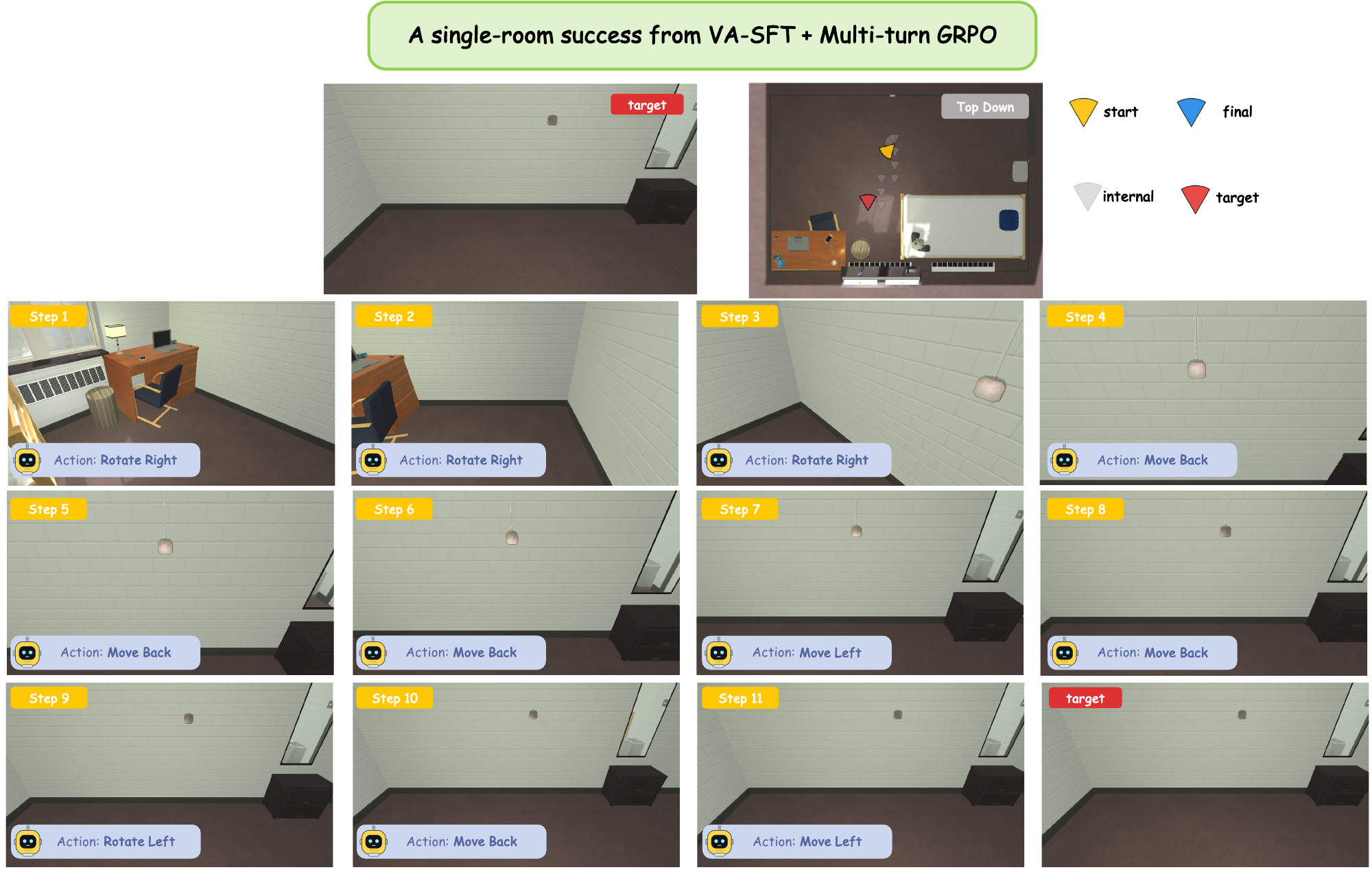}
\caption{\textbf{A single-room success from VA-SFT + Multi-turn GRPO.} On a single-room iTHOR scene, the policy translates and rotates to align with the target view within a handful of steps and terminates with Stop at the correct pose. Visual-action memory lets each step condition on the actual observation history, so the model no longer revisits previously tried yaws.}
\label{fig:qual-success-ithor}
\end{figure*}

\begin{figure*}[!t]
\centering
\includegraphics[width=0.97\textwidth]{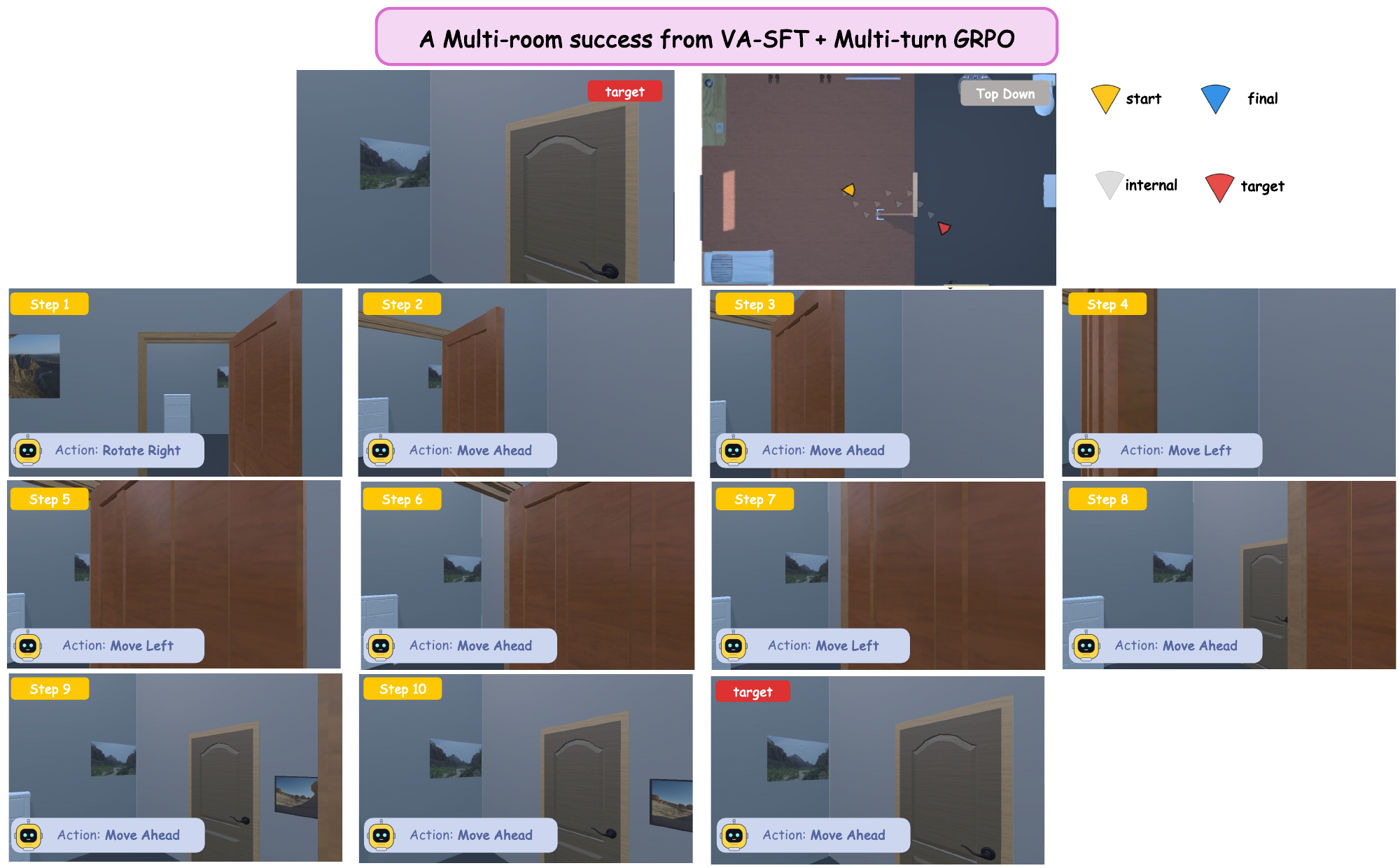}
\caption{\textbf{A multi-room success from VA-SFT + Multi-turn GRPO.} On a multi-room ProcTHOR scene, the policy traverses the layout across rooms and aligns with the target view before issuing Stop. Multi-room tasks are where the SFT initialisation alone is weakest (Table~\ref{tab:posttraining_main}); traces like this illustrate where Multi-turn GRPO adds its largest gain over VA-SFT.}
\label{fig:qual-success-procthor}
\end{figure*}

\end{document}